%% file: acml24.tex
\documentclass[pmlr]{jmlr}

\usepackage{longtable}

\usepackage{booktabs}
\usepackage[load-configurations=version-1]{siunitx} 
\usepackage{enumitem}
\usepackage{amsmath}
\usepackage{listings}
\usepackage{algpseudocode}
\usepackage{pifont}
\newcommand{\cmark}{\ding{51}}%
\newcommand{\xmark}{\ding{55}}%

\usepackage{lineno}
\usepackage{tikz}
\usetikzlibrary{patterns}
\usetikzlibrary{shadows.blur}

\newlist{inlineenum}{enumerate*}{1}
\setlist[inlineenum,1]{label=\arabic*), before=\hspace{.1em}, itemjoin={{, }}, itemjoin*={{, and }}}

\makeatletter
\let\Ginclude@graphics\@org@Ginclude@graphics 
\makeatother

\jmlrvolume{260}
\jmlryear{2024}
\jmlrworkshop{ACML 2024}
\jmlrpages{-}
\renewcommand{\thepage}{}

\title[Pic@Point: Cross-Modal Learning by Local and Global Point-Picture Correspondence]{Pic@Point: Cross-Modal Learning by Local and Global Point-Picture Correspondence}

  \author{\Name{Vencia Herzog} \Email{vencia.herzog@renumics.com}\\
  \addr Karlsruhe Institute of Technology\\
  \addr Renumics GmbH, Karlsruhe, Germany
  \AND
  \Name{Stefan Suwelack} \Email{stefan.suwelack@renumics.com}\\
 \addr Renumics GmbH, Karlsruhe, Germany
 }

\editors{Vu Nguyen and Hsuan-Tien Lin}

\begin{document}

\maketitle

\begin{abstract}
Self-supervised pre-training has achieved remarkable success in NLP and 2D vision.
However, these advances have yet to translate to 3D data.
Techniques like masked reconstruction face inherent challenges on unstructured point clouds, while many contrastive learning tasks lack in complexity and informative value.
In this paper, we present \textit{Pic@Point}, an effective contrastive learning method based on structural 2D-3D correspondences.
We leverage image cues rich in semantic and contextual knowledge to provide a guiding signal for point cloud representations at various abstraction levels.
Our lightweight approach outperforms state-of-the-art pre-training methods on several 3D benchmarks.
\end{abstract}

\begin{keywords}
self-supervised pre-training; cross-modal; 2D-3D correspondence; point clouds
\end{keywords}

\section{Introduction}
Point clouds are the preferred 3D representation for many applications, including autonomous driving, robotics, AR/VR, and various sensor technologies (LIDAR, SFM, Kinect Structured Light, etc).
However, the annotation of point clouds is associated with high costs, and labeled 3D scans are scarce.
This poses an obstacle to the development of robust, scalable deep learning models for point cloud analysis. 

Recent progress in self-supervised learning has led to significant advances in the areas of image recognition and natural language processing (NLP).
By using parts of the input data itself as a guiding signal, self-supervised learning bypasses the annotation bottleneck associated with training large neural networks.
This strategy reflects a shift from traditional task-specific training towards pre-training  general-purpose representations that are applicable across a wide range of tasks.
Extending these advances to 3D data remains challenging due to factors such as irregular information density and the unordered nature of point clouds. 

Generative modeling has seen a number of adaptations to 3D data using transformer-style architectures \citep{yu2022point,pang2022masked,zhang2022point}.
However, these approaches have yet to replace traditional architecture-based methods \citep{ma2022rethinking,qian2022pointnext}.
Two factors that contribute to the underperformance on point clouds are the lack of inductive biases in Standard Transformers and the use of point set similarity metrics (e.g., Chamfer Distance) in reconstruction, which are imprecise and hard to optimize \citep{wu2021density, huang2023learning}.

Contrastive learning approaches, on the other hand, typically aim at learning invariances to transformations or augmentations.
They have been shown to be highly competitive to generative modeling \citep{oord2018representation,chen2020simple}.
However, the quality of learned representations is highly dependent on the complexity and informative value of the contrastive task \citep{goyal2019scaling}, and many contrastive models only consider global views, neglecting local relationships \citep{qi2023contrast}.

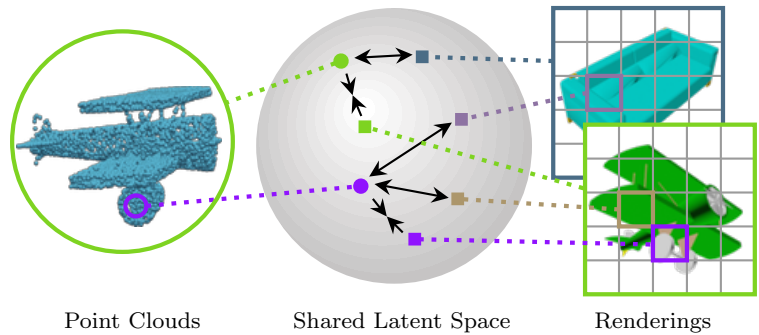
\begin{figure}[tb]
	\centering
	\resizebox{0.7\linewidth}{!}{%
		\input{tikz/illustration.tex}}
	\caption{\textbf{Illustration of the proposed structural 2D-3D correspondence.}
	Point cloud and image features of various scales are projected into a shared, representation-invariant latent space, where cross-modal correspondence is enforced.}
	\label{fig:illustration}
\end{figure}

\begin{figure}[htb]
	\centering
	\includegraphics[width=0.6\linewidth]{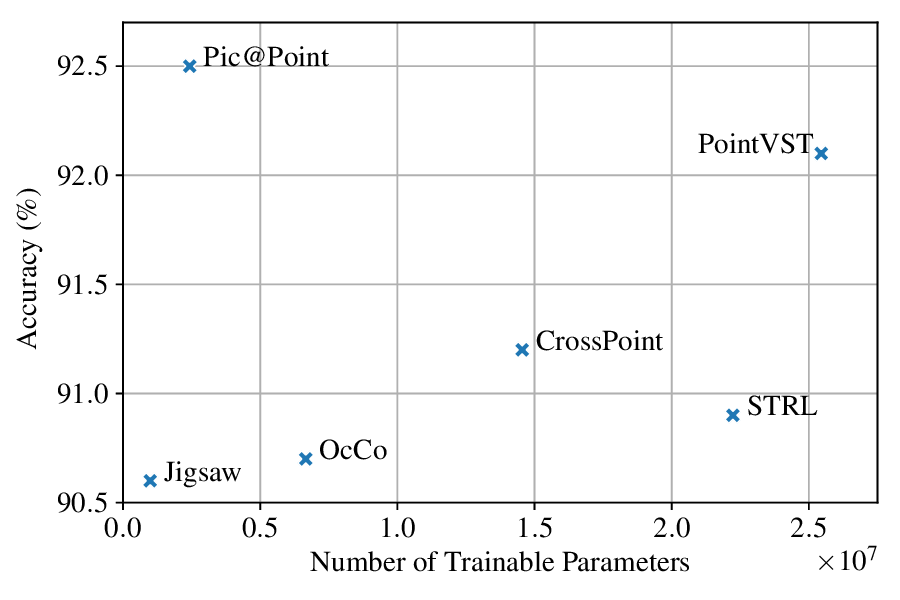}
	\caption{\textbf{Size of Pre-training Model vs. Linear SVM Accuracy on ModelNet40.} 
	Reported is the number of trainable parameters of self-supervised pre-training models with DGCNN backbone.}
	\label{fig:model-size}
\end{figure}

To address these limitations, we propose leveraging 2D-3D correspondences for contrastive representation learning on point clouds.
We present the \textbf{Pic@Point} model, which aims to learn point cloud representations by exploiting structural features of \textbf{\textit{pic}}tures \textbf{\textit{at}} various \textbf{\textit{point}} cloud abstraction levels.
Specifically, we extract 3D and 2D features at both global and local scales using a generic 3D backbone and a pre-trained 2D backbone, respectively. 
We then employ global and local, pose-conditioned projection heads to project these features into a common, representation-invariant latent space, as illustrated in \autoref{fig:illustration}.
This structural 2D-3D contrastive learning approach offers several advantages over existing methods:
\begin{itemize}
\item We effectively leverage features from pre-trained vision foundation models. 
In contrast, generative cross-modal methods  \citep{zhang2023pointvst,wang2023take} generate images from input point clouds using a custom 2D generator atop a 3D backbone, as opposed to integrating a powerful vision model.
\item In addition, our method is very lightweight while being highly effective, as it uses no decoder and employs a frozen 2D backbone. 
\autoref{fig:model-size} shows the size of different pre-training models in relation to linear accuracy on ModelNet40, showcasing the efficiency of our approach. 
\item Unlike existing contrastive methods \citep{afham2022crosspoint,wu2023self} which only learn global shape correspondences, Pic@Point provides guidance on a structural level.
By exploiting local correspondences, it provides pose-aware, positional guidance while benefiting from a larger number of contrastive samples. 
\end{itemize}

\section{Related Work}

\paragraph{Point Cloud Analysis}
Early efforts to apply neural networks to points clouds involved converting the input into structured forms such as voxel grids \citep{wu20153d,maturana2015voxnet} or multi-view images \citep{su2015multi}.
Sparse voxel-based methods \citep{riegler2017octnet,choy20194d} remain a prevalent approach for large-scale scene analysis.

With the introduction of PointNet \citep{qi2017pointnet}, point-based methods began to  apply neural architectures directly to raw point sets without any conversion or preprocessing steps.
A key concern is ensuring permutation invariance, often addressed by aggregating information using symmetric functions.  
Deep set architectures can be divided into three categories: MLP-methods \citep{qi2017pointnet,qi2017pointnet++,wang2019dynamic,ma2022rethinking}, convolution-based methods \citep{li2018pointcnn,thomas2019kpconv}, and attention-based methods \citep{yu2022point,pang2022masked,zhang2022point}.

\paragraph{Self-supervised Point Cloud Pre-training}
Pre-training is a widely used transfer learning approach in which a model is initially trained on a \textit{pretext} task and subsequently fine-tuned on a \textit{downstream} task.
Self-supervised pre-training does not use any labels during the pre-training phase, which enables a wide scope of available pre-training data.

Pretext tasks are typically either generative or contrastive.
A common generative approach is autoencoding, where the input is recovered under some form of corruption \citep{chen2021shape} or masking \citep{wang2021unsupervised}.
Transformer architectures \citep{devlin2018bert,he2022masked} combine masked reconstruction with multi-head self-attention mechanisms. 
Point-BERT \citep{yu2022point} and Point-MAE \citep{pang2022masked} adapt language and vision Transformers to point clouds.
Generative models typically reconstruct directly in point cloud space, which is computationally expensive and hard to optimize \citep{wu2021density, huang2023learning}.

Contrastive learning uses available pairs of similar and dissimilar data points to learn an embedding space where the distance between data points reflects a measure of their similarity.
This is done by contrasting a sample with augmented versions \citep{he2020momentum,chen2020simple} or by capturing the relationship between local features and their global context \citep{oord2018representation,hjelm2018learning}.

Contrastive methods benefit from diverse, informative data to mitigate overfitting issues and counterbalance representation deficits.
To this end, we propose a cross-modal method leveraging image cues rich in structural and semantic context.

\paragraph{Cross-Modal 3D Representation Learning}
The potential of joint 2D-3D learning has been recognized by previous works.
One line of work adapts pre-trained vision and language models for 3D point cloud analysis using strategies such as prompt tuning \citep{zhang2022pointclip,wang2022p2p}, 2D-to-3D architectural modification \citep{xu2022image2point, qian2022pix4point} and knowledge distillation \citep{dong2022autoencoders,qi2023contrast} techniques.
These methods cannot be directly applied to generic 3D backbones because they require specialized architectures or adaptation processes, often having large memory requirements and difficulties with 3D extensibility.

A different approach is to design cross-modal pretext tasks directly on 3D models, allowing the use of generic 3D backbones for downstream tasks. 
PointVST \citep{zhang2023pointvst} and TAP \citep{wang2023take} employ a generative cross-modal approach that generates images from point clouds at specified camera views, but they lack knowledge integration from pre-trained vision models.
CrossPoint \citep{afham2022crosspoint} performs cross-modal and inter-modal contrastive learning to align point clouds with 2D renderings and with augmented versions.
They leverage only global correspondences, resulting in a coarse guiding signal.
Conversely, point-pixel level correspondence methods such as \cite{tran2022self} require costly upsampling layers and loss computation, while being unnecessarily fine-grained for learning meaningful contextual relationships.

\begin{figure*}[htb]
	\centering
	\resizebox{\linewidth}{!}{%
	\input{tikz/architecture.tex}}
	\caption{\textbf{Overview of the proposed Pic@Point workflow.}
	We extract top-level local and aggregated global point cloud and image features using a generic 3D backbone and a frozen 2D backbone.
	Subsequently, the features are processed through global and local, pose-conditioned projection heads $\mathcal{H}$.}
	\label{fig:architecture}
\end{figure*}
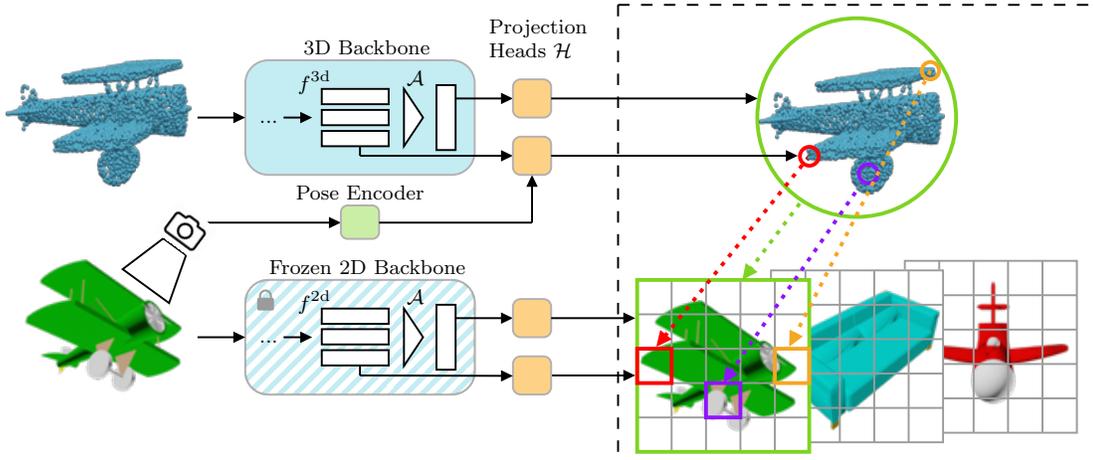

\section{Proposed Method}
We propose structural 2D-3D correspondence learning for self-supervised pre-training of point cloud representations.
The importance of incorporating structural knowledge has been demonstrated in prior uni-modal research \citep{hjelm2018learning,oord2018representation,rao2020global}.
Our novel cross-modal approach enriches point cloud information with structured, semantic image cues to provide a comprehensive guiding signal for 3D understanding.

\subsection{Overview}
Let $\mathcal{D} = \{(P_i,I_i)\}_{i=1}^{|\mathcal{D}|}$ denote an unlabeled dataset of point clouds $P_i \in \mathbb{R}^{N \times 3}$, where $N$ is the number of points, and shape renderings $I_i \in \mathbb{R}^{H \times W \times 3}$ of size $H \times W$.
Rendering $I_i$ is captured from a random camera view point with view matrix $M_{view} \in \mathbb{R}^{3 \times 3}$ and projection matrix $M_{proj} \in \mathbb{R}^{3 \times 3}$.
\autoref{fig:architecture} depicts the overall architecture of our proposed Pic@Point model.
It consists of the following modules: 
\begin{inlineenum}
	\item a \textit{3D Backbone} extracts local and global point cloud features, obtained as top-level positional features $f^\text{3d}$ and final shape embeddings $\mathcal{A}(f^\text{3d})$, after a global pooling function $\mathcal{A}$
	\item a frozen \textit{2D Backbone} returns top-level local and global image features as extracted by a pre-trained vision model (e.g., ResNet, ViT), denoted as $f^\text{2d}$ and $\mathcal{A}(f^\text{2d})$ 
	\item the \textit{Projection Heads $\mathcal{H}$} project the point cloud and image features into a shared, representation-invariant latent space $\chi \subseteq \mathbb{R}^{d}$.
\end{inlineenum}

\subsection{Projection into Shared Latent Space}
To unify knowledge and produce rich, transferable representations, we project the modal features into a shared latent space, where a contrastive loss is applied between projected features within the mini-batch.
The global projection heads $\mathcal{H}^\text{3d}_\text{glb}, \mathcal{H}^\text{2d}_\text{glb}$ consist of simple multi-layer perceptrons (MLPs) with two layers.
The local projection heads $\mathcal{H}_\text{lcl}^\text{3d}$, $\mathcal{H}_\text{lcl}^\text{2d}$ use Conv1d and Conv2d layers with kernel size $1$ and stride $1$, respectively, to apply transformations in the channel dimension while preserving the spatial dimensions.
All projected features are L2-normalized.

\paragraph{Pose Encoding}
We facilitate spatially aware correspondence between local point cloud and image features by integrating a pose encoding into the local point cloud projection head $\mathcal{H}_\text{lcl}^\text{3d}$.
This is necessary because while it is assumed that a complete shape can be uniquely identified in a representation of any modality, this may not hold true for a local shape region.
For instance, symmetric features like the red circled plane wing tip depicted in  \autoref{fig:illustration} could potentially belong to any of the image snippets depicting a plane wing tip, if no additional pose information is given.
The pose encoding is calculated via a two-layer MLP that transforms $M_{view}$ into a $64$-dimensional vector that is concatenated to the output of the first convolutional layer in $\mathcal{H}_\text{lcl}^\text{3d}$.

\subsection{Cross-Modal Correspondence Task}

\paragraph{Point-to-Pixel Mapping}
For the local cross-modal correspondence task, we begin by establishing the ground truth mapping from points to image positions using projective transformations with $M_{view},M_{proj} \in \mathbb{R}^{4 \times 4}$.
Given a local point cloud embedding $z_l$, let $(u,v) \in [0,1]^2$ denote the image position projected from its center point. 
With top-level downsampled image regions of size $7 \times 7$, such as produced by ResNet \citep{he2016deep}, the indexed position $(i,j)$ of the corresponding image region embedding $q_{ij}$ is determined as $(i,j)=\lfloor(u \cdot 7,v \cdot 7)\rfloor$.

Subsequently, a contrastive learning objective can be applied to the cross-modal correspondences.
For each of the $L$ local point cloud region embeddings $\{z_l\}_{l = 1, ..., L}$, we  pull close the corresponding image region embedding $q^+$ of the same object, while pushing away all other image region embeddings $\bigl\{q_{ij}^k \neq q^+\bigr\}^{k = 1, ..., m}_{i,j = 1, ..., 7}$ within the mini-batch of size $m$.
To achieve this, we apply the InfoNCE loss function \citep{oord2018representation} with temperature hyper-parameter $\tau$: 

\begin{align}
\mathcal{L}_\text{lcl} &= \frac{1}{L} \sum_l \mathcal{L}_\text{lcl}^l, \qquad \text{with} \\
\mathcal{L}_\text{lcl}^l &= -\log\frac{\exp(z_l^\top q^+/\tau)}{\sum_{i,j,k} \exp(z_l^\top q_{ij}^k/\tau)}.
\end{align}

Similarly, for global point cloud embedding $z$, we pull close the global image embedding $q^+$ of the same object, while pushing away all other global image embeddings $\bigl\{q^k \neq q^+\bigr\}^{k = 1, ..., m}$,

\begin{equation}
\mathcal{L}_\text{glb} = -\log\frac{\exp(z^\top q^+/\tau)}{\sum_{k} \exp(z^\top q^k/\tau)}.
\end{equation}

The overall loss function is given by $\mathcal{L} = \mathcal{L}_\text{lcl} + \mathcal{L}_\text{glb}$.

\section{Experiments}
In the following, we present our experimental setups and results of Pic@Point pre-training using different point cloud backbones.
We conduct experiments on four standard benchmarks: \textit{ModelNet40} \citep{wu20153d} and \textit{ScanObjectNN} \citep{uy2019revisiting} for object classification, \textit{ShapeNetPart} \citep{yi2016scalable} for part segmentation, and \textit{S3DIS} \citep{armeni20163d} for semantic scene segmentation.

\subsection{Pre-Training Setup}

\paragraph{Pre-training Dataset}
Following common practice, we pre-train on the ShapeNet \citep{chang2015shapenet} dataset, which consists of more than \num{50000} CAD models from 55 semantic categories.
We obtain point clouds by randomly sampling \num{1024} points from each object.
Furthermore, we obtain renderings of size $224 \times 224$ from $20$ different view points placed in a regular dodecahedron around the object, saving projection and camera matrices $M_\text{proj}, M_\text{view}$.
We randomly select a single rendering per object at each iteration and apply random rotation as augmentation.

\paragraph{Architectures}
We conduct experiments using two prominent point cloud backbones: DGCNN~\citep{wang2019dynamic} and PointNeXt~\citep{qian2022pointnext}.
DGCNN is a standard architecture used for benchmarking many existing self-supervised methods.
PointNeXt is a more recent hierarchical model achieving state-of-the-art results across many 3D tasks.
Besides drawing comparison to related contrastive and generative cross-modal methods using the same backbones, we also compare against recent transformer-style methods.
For the image backbone we use a light-weight ResNet-18 \citep{he2016deep}.

During the pre-training phase, the global and local point cloud features are extracted directly from the 3D backbone.
Subsequently, the features are projected to shared latent space with dimension $d=512$.
After pre-training, the projection heads are dropped, and the 3D backbone is used exclusively.

\paragraph{Implementation Details}
The experiments are implemented in PyTorch using the OpenPoints framework  \citep{qian2022pointnext}.
We utilize the Adam optimizer with CosineAnnealing and a weight decay of $1e^{-6}$ for pre-training.
For the point cloud branch we use an initial learning rate of $1e^{-3}$, for the image branch we set a lower learning rate of $5e^{-5}$.
We train with a batch size of $32$.
For the downstream tasks, we follow the training and evaluation settings of \cite{qian2022pointnext}.

\subsection{Downstream Tasks}
In the following, we present experimental results on object classification, part segmentation and scene segmentation.

\begin{table}[htb]
	\sisetup{detect-weight=true,detect-inline-weight=math,separate-uncertainty=true,multi-part-units=single}
	\centering
	\begin{tabular}{lS[table-format=2.1]S[table-format=2.1]}
		\toprule
		Method & {ModelNet40} & {ScanObjectNN}\\
		\midrule
		{FoldingNet~\citep{yang2018foldingnet}}  & 88.4 & {-} \\
		{VIP-GAN~\citep{han2019view}} &  90.2 & {-} \\
		{Point-BERT~\citep{yu2022point}} &  87.4 & {-} \\
		{Point-MAE~\citep{pang2022masked}} &  91.0 & 77.7 \\
		{Point-M2AE~\citep{zhang2022point}} & 92.9 & 84.1 \\
		\midrule
		{DGCNN+Jigsaw~\citep{sauder2019self}} &  90.6 & 59.5 \\
		{DGCNN+STRL}~\citep{huang2021spatio} & 90.9 & 77.9 \\
		{DGCNN+OcCo~\citep{wang2021unsupervised}} &  90.7 & 78.3 \\ 
		{DGCNN+CrossPoint~\citep{afham2022crosspoint}} &  91.2 & 81.7 \\
		{DGCNN+PointVST~\citep{zhang2023pointvst}} &  92.1 & {-} \vspace{0.15cm}\\
		{DGCNN+Pic@Point (ours)} & $\mathbf{92.5}$ & $\mathbf{85.7}$ \\
		{DGCNN+Pic@Point (ours, w/ normals)} &  $\mathbf{92.9}$ & {-} \\
		\bottomrule	
	\end{tabular}
	\caption{\textbf{Linear SVM Classification on ModelNet40 and ScanObjectNN (OBJ\_BG).}
		We compare against methods using a specialized point cloud architecture (\textit{top}), and methods using a DGCNN backbone (\textit{bottom}).
		We report the overall accuracy (\%) with $1024$ points.} 
	\label{fig:svm-classification}
\end{table}

\subsubsection{Object Classification}\label{sec:classification}
To evaluate the effectiveness of our proposed Pic@Point method for object classification, we conduct experiments on the synthetic dataset ModelNet40 \citep{wu20153d} and the real-world scanned dataset ScanObjectNN \citep{uy2019revisiting}.
ModelNet40 consists of \num{12331} 3D CAD models from 40 object categories.
ScanObjectNN contains 15 categories of real-world indoor scans with \num{2903} unique  object instances.
We evaluate on all three common variants of the the ScanObjectNN dataset: OBJ\_BG contains complete object scans, OBJ\_ONLY uses background cropping, and PB\_T50\_RS contains perturbed versions of the scans.
We sample $1024$ points on each object for both training and testing.

We use two transfer learning protocols to evaluate the effectiveness of our proposed Pic@Point model: linear probing with an SVM and fine-tuning on the downstream dataset.

\paragraph{Linear SVM Results}
We test the representation capability of Pic@Point by fitting a linear SVM on the features extracted from the pre-trained point cloud backbone.
In \autoref{fig:svm-classification} we report our linear classification results on ModelNet40 and ScanObjectNN (OBJ\_BG).
The upper part of the table shows existing self-supervised methods employing specialized point cloud architectures, including transformer-style methods.
These methods focus primarily on point cloud encoding architectures rather than on the design of pretext tasks.
The lower part of the table shows architecture-agnostic pre-training methods, which are compared using a DGCNN backbone.

Pic@Point significantly outperforms competing methods on a DGCNN backbone with $92.5\%$ and $85.7\%$ linear classification accuracy on ModelNet40 and ScanObjectNN (OBJ\_BG), respectively.
By incorporating normals as additional input, we further improve to $92.9\%$ accuracy on ModelNet40.
We achieve margins of $+0.4\%$ and $+4.0\%$ over the second best methods PointVST \citep{zhang2023pointvst} and CrossPoint \citep{afham2022crosspoint}, respectively.
This underscores the effectiveness of our proposed structural 2D-3D correspondence learning over existing cross-modal approaches such as PointVST, a generative method, and CrossPoint, a contrastive method that relies solely on global correspondences.

Notably, the improvements over existing methods are more pronounced on ScanObjectNN than on ModelNet40.
This may be attributed to Pic@Point's enhanced generalization ability through learning modality-invariant features, making it more robust on real-world data with irregular sampling and noise.
We outperform methods employing larger transformer-style architectures on ScanObjectNN, achieving a margin of $+1.6\%$ over the second best method Point-M2AE \citep{zhang2022point}, which uses a large multi-scale Transformer architecture.

\begin{table*}[htb]
	\sisetup{detect-weight=true,detect-inline-weight=math,separate-uncertainty=true,multi-part-units=single}
	\centering
	\resizebox{\linewidth}{!}{%
		\begin{tabular}{lS[table-format=2.1]lll}
			\toprule
			Method & {\#Params (M)} & {OBJ\_BG} & {OBJ\_ONLY} &{PB\_T50\_RS} \\
			\midrule
			{PointNet~\citep{qi2017pointnet}} & 3.5 & 73.3 & 79.2 & 68.2 \\
			{DGCNN~\citep{wang2019dynamic}} & 1.8 & 82.8 & 86.2 & 78.1 \\
			{PointNeXt-S~\citep{qian2022pointnext}} & 1.4 & {-} & {-} & 87.7 \\
			{PointMLP~\citep{ma2022rethinking}} & 12.6 & {-} & {-} & 85.4 \\
			\midrule
			{Pix4Point~\citep{qian2022pix4point}} & 22.1 & {-} & {-} & 87.9 \\
			{ACT~\citep{dong2022autoencoders}} & 22.1 & 93.3 & 91.9 & 88.2 \\
			{P2P~\citep{wang2022p2p}} & 195.8 & {-} & {-} & 89.3 \\
			{I2P-MAE~\citep{zhang2023learning}} & 12.9 & 94.2 & 91.6 & 90.1 \\
			{ReCon~\citep{qi2023contrast}} & 43.6 & 95.2 & 93.6 & 90.6 \\
			\midrule
			{[T]ransformer~\citep{vaswani2017attention}} & 22.1 & 79.9 & 80.6 & 77.2 \\
			{[T]+OcCo~\citep{yu2022point}} & 22.1 & 84.9 & 85.5 & 78.8 \\
			{Point-BERT~\citep{yu2022point}} & 22.1 & 87.4 & 88.1 & 83.1 \\ 
			{Point-MAE~\citep{pang2022masked}} & 22.1 & 90.0 & 88.3 & 85.2 \\ 
			{Point-M2AE~\citep{zhang2022point}} & 15.3 & 91.2 & 88.8 & 86.4 \\ 
						{[T]+PointVST~\citep{zhang2023pointvst}} & 22.1 & {-} & {-} & 86.6 \\
			{[T]+TAP~\citep{wang2023take}} & 22.1 & 90.4 & 89.5 & 85.7 \\
			{PointMLP+TAP~\citep{wang2023take}} & 12.6 & {-} & {-} & 88.5 \\
			\midrule
			{PointNeXt-S (reproduce)} & 1.4 & 91.2 & 89.9 & 87.2 \\
			{$\quad +$ Pic@Point} & 1.4 & $\mathbf{94.0}$ $\color{blue} (+2.8)$ & $\mathbf{92.6}$ $\color{blue} (+2.7)$ & $\mathbf{88.1}$ $\color{blue} (+0.9)$ \\
			\bottomrule	
	\end{tabular}}
	\caption{\textbf{Real-world 3D Classification on ScanObjectNN Variants.} 
		We report the number of inference model parameters (M) and the overall accuracy (\%) with $1024$ points.}
	\label{fig:finetuning-classification}
\end{table*}

\paragraph{Fine-tuning Results}
Next, we perform extensive fine-tuning experiments on all three variants of the  ScanObjectNN dataset, which has established itself as a favored classification benchmark \citep{qian2022pointnext, ma2022rethinking}.
The results are shown in \autoref{fig:finetuning-classification}.
The top part of the table shows supervised 3D models trained from scratch.
The middle part shows 2D-to-3D methods that use large-scale vision foundation models with specialized architectures and 3D downstream adaptations such as visual prompt tuning or multi-stage knowledge distillation.
We note that our model does not directly compete with these methods due to their use of specialized architectures with large memory requirements and 2D-to-3D adaptations.
Nevertheless, we include them to provide a comprehensive overview of related research.
The lower part of the table lists self-supervised methods employing state-of-the-art 3D point cloud models, many of which are based on Transformer architectures.

We perform experiments using the PointNeXt-S \citep{qian2022pointnext} backbone.
We observe significant improvements compared to training from scratch: $+2.8\%$ on OBJ\_BG, $+2.7\%$ on OBJ\_ONLY and $+0.9\%$ on PB\_T50\_RS.
Pic@Point surpasses all self-supervised methods with transformer-style point cloud encoders on all three variants of ScanObjectNN, while having a factor $\times \frac{1}{10}$ smaller model size.
On PB\_T50\_RS, Pic@Point is outperformed only by TAP \citep{wang2023take} using a PointMLP backbone.
On OBJ\_BG and OBJ\_ONLY, Pic@Point outperforms all competing 3D methods.

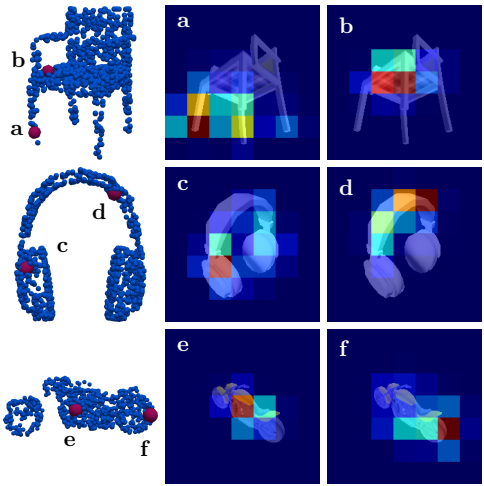
\begin{figure}[htb]
	\centering
	\resizebox{0.45\linewidth}{!}{%
		\input{tikz/heatmap.tex}}
	\caption{\textbf{Visualization of local 2D-3D Correspondences}.
	The first column displays point clouds with two example points highlighted.
	The second and third columns display heatmaps of the feature distances between one of these points and the image patches of the corresponding object in latent space.}
	\label{fig:heatmap-vis}
\end{figure}

\paragraph{Visualization}
In \autoref{fig:heatmap-vis} we show visualizations of local 2D-3D correspondences learned during Pic@Point pre-training.
For each object, the correspondences between the object's image patches and two exemplar points are depicted.
Two key observations can be made:
Firstly, Pic@Point successfully learns to correlate structures of point cloud regions with image patches depicting semantically similar features.
Secondly, it accurately matches corresponding image features by incorporating pose information.

\begin{table}[htb]
	\sisetup{detect-weight=true,detect-inline-weight=math,separate-uncertainty=true,multi-part-units=single}
	\centering
		\resizebox{0.5\linewidth}{!}{%
	\begin{tabular}{cc|c|c|l}
		\toprule
		$\mathcal{L}_\text{lcl}$ & $\mathcal{L}_\text{glb}$ & Pose Cond. & Normals & \text{OA (\%)} \\
		\midrule
		\xmark & \cmark & \cmark & \cmark & $92.3$ $\color{blue} (-0.2)$ \\
		\cmark & \xmark & \cmark & \cmark & $92.3$ $\color{blue} (-0.2)$ \\
		\cmark & \cmark & \xmark & \cmark & $91.8$ $\color{blue} (-0.7)$ \\
		\cmark & \cmark & \cmark & \cmark & $92.8$ $\color{blue} (+0.3)$ \\
		\midrule
		\cmark & \cmark & \cmark & \xmark & $92.5$ \\
		\bottomrule
	\end{tabular}}
	\caption{\textbf{Ablation Studies.}
	Contributions of local and global correspondence, the impact of pose conditioning, and the use of point normals are evaluated via linear SVM classification on ModelNet40 using a DGCNN backbone.
	}
	\label{tab:ablation}
\end{table}

\paragraph{Ablation Studies}
\autoref{tab:ablation} presents ablation studies on key components.
First, we evaluate the local and global correspondence losses $\mathcal{L_\text{lcl}}$ and $\mathcal{L_\text{gbl}}$, both of which have equal impact on the results.
Second, we confirm that the pose encoding aids in learning from local correspondences.
A drop of $-0.7\%$ accuracy is observed when pose information is omitted.
Lastly, while our standard Pic@Point model excludes normals to ensure fair comparison with related methods, we show that incorporating normal information during pre-training and downstream fine-tuning, when available, provides significant benefits.

\begin{table}[htb]
	\sisetup{detect-weight=true,detect-inline-weight=math,separate-uncertainty=true,multi-part-units=single}
	\centering
	\begin{tabular}{lS[table-format=2.1]S[table-format=2.1]}
		\toprule
		Method & {ins. mIoU (\%)}  & {cls. mIoU (\%)}\\
		\midrule
		{PointNet~\citep{qi2017pointnet}} & 83.7 & 80.4 \\
		{DGCNN~\citep{wang2019dynamic}} & 85.2 & 82.3 \\
		{PointNeXt-S~\citep{qian2022pointnext}} & 86.7 & 84.4 \\
		{PointMLP~\citep{ma2022rethinking}} & 86.1 & 84.6\\
		\midrule
		{Transformer~\citep{vaswani2017attention}} & 85.1 & 83.4 \\
		{Point-BERT~\citep{yu2022point}} &  85.6 & 84.1 \\
		{Point-MAE~\citep{pang2022masked}} & 86.1 & 84.2 \\
		\midrule
		{DGCNN+Jigsaw~\citep{sauder2019self}}  & 85.3 & 82.3 \\
		{DGCNN+OcCo~\citep{wang2021unsupervised}} & 85.0 &  {-} \\
		{DGCNN+CrossPoint~\citep{afham2022crosspoint}} & 85.5 & {-} \\
		{DGCNN+PointVST~\citep{zhang2023pointvst}} & $\mathbf{87.4}$ & {-} \vspace{0.15cm} \\
		{DGCNN+Pic@Point (ours)} & 85.8 & $\mathbf{83.0}$ \\
		\bottomrule	
	\end{tabular}
	\caption{\textbf{Part Segmentation on ShapeNetPart.}
		We report the mean IoU across all instances (ins.) and across all classes (cls.) with $2048$ points.}
	\label{tab:part-segmentation}
\end{table}

\subsubsection{Part Segmentation}\label{sec:part-segmentation}
ShapeNetPart \citep{yi2016scalable} is a widely used dataset for object part segmentation.
It contains \num{16881} pre-aligned CAD models from 16 object classes and has a total of 50 part categories.
We use the same pre-training setup as for classification, pre-training on $1024$ randomly sampled points per object on the ShapeNet dataset.
For downstream fine-tuning on ShapeNetPart, we train and test on $2048$ points.
\autoref{tab:part-segmentation} reports the mean intersection over union (mIoU) averaged over all instances (ins.) and averaged over all object classes (cls.) with DGCNN backbone.

Overall, ShapeNetPart results are less distinguishable compared to other benchmarks.
Both Jigsaw \citep{sauder2019self} and OcCo \citep{wang2021unsupervised} show no significant improvement over training from scratch using DGCNN.
Pic@Point demonstrates slightly better performance than previous pre-training methods on a DGCNN backbone, except for PointVST \citep{zhang2023pointvst}, which reports an exceptional result of $87.4\%$.
Compared to CrossPoint \citep{afham2022crosspoint}, which uses  global 2D correspondences, Pic@Point shows an improvement of $+0.3\%$.

\begin{table}[htb]
	\sisetup{detect-weight=true,detect-inline-weight=math,separate-uncertainty=true,multi-part-units=single}
	\centering
	\begin{tabular}{lll}
		\toprule
		Method & {mIoU (\%)} & {mAcc (\%)}\\
		\midrule
		{PointNet~\citep{qi2017pointnet}} & 41.1 &  49.0 \\
		{DGCNN~\citep{wang2019dynamic}} & 47.9 & {-} \\
		{PointNeXt-S~\citep{qian2022pointnext}} & 63.4 & {-}\\
		{PointNeXt-XL~\citep{qian2022pointnext}} & 70.5 & {-}\\
		\midrule
		{Transformer~\citep{vaswani2017attention}} &  60.0 & 68.6 \\ 
		{Point-BERT~\citep{yu2022point}} &  60.8 &  69.9 \\
		{Point-MAE~\citep{pang2022masked}} &  60.8 &  69.9\\
		\midrule
		{PointNeXt-S (reproduce)} & 62.9 & 69.5 \\
		{$\quad +$ Pic@Point} & $\mathbf{63.6}$ $\color{blue} (+0.7)$ &  $\mathbf{71.1}$ $\color{blue} (+0.6)$ \\
		\bottomrule	
	\end{tabular}
	\caption{\textbf{Scene-Level Semantic Segmentation on S3DIS Area 5.}
		We report mean IoU and mean accuracy.}
	\label{tab:scene-segmentation}
\end{table}

\subsubsection{Scene Segmentation}\label{sec:scene-segmentation}
Semantic segmentation on large 3D scenes challenges the understanding of contextual relationships and coherent semantic interpretation.
S3DIS \citep{armeni20163d} is a scene segmentation benchmark consisting of $6$ types of large scanned indoor areas with 13 semantic categories.
Following common practice, we test on the largest area, Area 5, and fine-tune on the remaining areas.
For training, the point clouds are downsampled with a voxel size of $0.04$m and sub-sampled to \num{24000} points.
Testing is conducted on the entire scene.

Pic@Point consistently improves performance over training from scratch, increasing mIoU by $+0.7\%$ and mAcc by $+0.6\%$.
It outperforms leading transformer-style methods by a margin of $+2.8\%$ mIoU.
The achieved scene segmentation results indicate that Pic@Point is superior in performing dense prediction tasks through its integration of rich semantic cues and structural, pose-aware guidance.

\section{Conclusions}
This paper presents Pic@Point, a self-supervised pre-training method that leverages 2D-3D correspondences at local and global scales.
Our proposed method uses a simple contrastive learning framework to integrate point cloud and image features across various abstraction levels, providing a guiding signal that is rich in semantic and structural knowledge.
Pic@Point pre-training significantly outperforms existing self-supervised pre-training methods, including those based on Transformer architectures, in various 3D understanding tasks.
Its lightweight, architecture-agnostic design offers distinct practical advantages and
benefits from future advancements in point cloud technologies.

\acks{This work was supported by funding from the Federal Ministry of Education and Research (BMBF) through the research project DAVIS within the "Research, Development and Use of Artificial Intelligence Methods in SMEs" program.}

\bibliography{acml24}

%
%
%
%

\end{document}

%% file: tikz/illustration.tex
\graphicspath{{./figures/architecture/}} 


\tikzset {_u2difim6g/.code = {\pgfsetadditionalshadetransform{ \pgftransformshift{\pgfpoint{89.1 bp } { -108.9 bp }  }  \pgftransformscale{1.32 }  }}}
\pgfdeclareradialshading{_yesh8nr13}{\pgfpoint{-72bp}{88bp}}{rgb(0bp)=(1,1,1);
	rgb(0bp)=(1,1,1);
	rgb(25bp)=(0.7,0.69,0.69);
	rgb(400bp)=(0.7,0.69,0.69)}
\tikzset{every picture/.style={line width=0.75pt}} 

\begin{tikzpicture}[x=0.75pt,y=0.75pt,yscale=-1,xscale=1]

\draw  [color={rgb, 255:red, 255; green, 255; blue, 255 }  ,draw opacity=1 ][fill={rgb, 255:red, 255; green, 255; blue, 255 }  ,fill opacity=1 ] (350,325) -- (430,325) -- (430,405) -- (350,405) -- cycle ;
\draw (390.99,367) node  {\includegraphics[width=58.52pt,height=51pt]{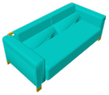}};
\draw  [color={rgb, 255:red, 155; green, 155; blue, 155 }  ,draw opacity=1 ] (350,325) -- (430,325) -- (430,405) -- (350,405) -- cycle ;
\draw [color={rgb, 255:red, 155; green, 155; blue, 155 }  ,draw opacity=1 ]   (366,325) -- (366,405) ;
\draw [color={rgb, 255:red, 155; green, 155; blue, 155 }  ,draw opacity=1 ]   (382,325) -- (382,405) ;
\draw [color={rgb, 255:red, 155; green, 155; blue, 155 }  ,draw opacity=1 ]   (398,325) -- (398,405) ;
\draw [color={rgb, 255:red, 155; green, 155; blue, 155 }  ,draw opacity=1 ]   (414,325) -- (414,405) ;
\draw [color={rgb, 255:red, 155; green, 155; blue, 155 }  ,draw opacity=1 ]   (350,341) -- (430,341) ;
\draw [color={rgb, 255:red, 155; green, 155; blue, 155 }  ,draw opacity=1 ]   (350,357) -- (430,357) ;
\draw [color={rgb, 255:red, 155; green, 155; blue, 155 }  ,draw opacity=1 ]   (350,373) -- (430,373) ;
\draw [color={rgb, 255:red, 155; green, 155; blue, 155 }  ,draw opacity=1 ]   (350,389) -- (430,389) ;

\draw  [color={rgb, 255:red, 74; green, 108; blue, 134 }  ,draw opacity=1 ][line width=1.5]  (350,325) -- (430,325) -- (430,405) -- (350,405) -- cycle ;
\draw (145.34,394.98) node  {\includegraphics[width=72.51pt,height=49.53pt]{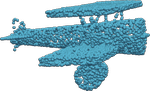}};
\draw  [color={rgb, 255:red, 144; green, 19; blue, 254 }  ,draw opacity=1 ][line width=1.5]  (148.4,418.33) .. controls (148.4,415.61) and (150.61,413.41) .. (153.33,413.41) .. controls (156.06,413.41) and (158.27,415.61) .. (158.27,418.33) .. controls (158.27,421.05) and (156.06,423.26) .. (153.33,423.26) .. controls (150.61,423.26) and (148.4,421.05) .. (148.4,418.33) -- cycle ;
\draw  [draw opacity=0][shading=_yesh8nr13,_u2difim6g] (210,390) .. controls (210,354.1) and (239.1,325) .. (275,325) .. controls (310.9,325) and (340,354.1) .. (340,390) .. controls (340,425.9) and (310.9,455) .. (275,455) .. controls (239.1,455) and (210,425.9) .. (210,390) -- cycle ;
\draw  [color={rgb, 255:red, 255; green, 255; blue, 255 }  ,draw opacity=1 ][fill={rgb, 255:red, 255; green, 255; blue, 255 }  ,fill opacity=1 ] (365,380) -- (445,380) -- (445,460) -- (365,460) -- cycle ;
\draw (402.71,420) node  {\includegraphics[width=56.57pt,height=48pt]{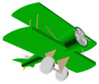}};
\draw  [color={rgb, 255:red, 155; green, 155; blue, 155 }  ,draw opacity=1 ] (365,380) -- (445,380) -- (445,460) -- (365,460) -- cycle ;
\draw [color={rgb, 255:red, 155; green, 155; blue, 155 }  ,draw opacity=1 ]   (381,380) -- (381,460) ;
\draw [color={rgb, 255:red, 155; green, 155; blue, 155 }  ,draw opacity=1 ]   (397,380) -- (397,460) ;
\draw [color={rgb, 255:red, 155; green, 155; blue, 155 }  ,draw opacity=1 ]   (413,380) -- (413,460) ;
\draw [color={rgb, 255:red, 155; green, 155; blue, 155 }  ,draw opacity=1 ]   (429,380) -- (429,460) ;
\draw [color={rgb, 255:red, 155; green, 155; blue, 155 }  ,draw opacity=1 ]   (365,396) -- (445,396) ;
\draw [color={rgb, 255:red, 155; green, 155; blue, 155 }  ,draw opacity=1 ]   (365,412) -- (445,412) ;
\draw [color={rgb, 255:red, 155; green, 155; blue, 155 }  ,draw opacity=1 ]   (365,428) -- (445,428) ;
\draw [color={rgb, 255:red, 155; green, 155; blue, 155 }  ,draw opacity=1 ]   (365,444) -- (445,444) ;
\draw  [color={rgb, 255:red, 126; green, 211; blue, 33 }  ,draw opacity=1 ][line width=1.5]  (365,380) -- (445,380) -- (445,460) -- (365,460) -- cycle ;
\draw  [color={rgb, 255:red, 172; green, 150; blue, 107 }  ,draw opacity=1 ][line width=1.5]  (381,412) -- (397,412) -- (397,428) -- (381,428) -- cycle ;
\draw  [color={rgb, 255:red, 144; green, 19; blue, 254 }  ,draw opacity=1 ][line width=1.5]  (397,428) -- (413,428) -- (413,444) -- (397,444) -- cycle ;
\draw [color={rgb, 255:red, 144; green, 19; blue, 254 }  ,draw opacity=1 ][line width=1.5]  [dash pattern={on 1.69pt off 2.76pt}]  (158.27,418.33) -- (259.33,409) ;
\draw [shift={(259.33,409)}, rotate = 354.72] [color={rgb, 255:red, 144; green, 19; blue, 254 }  ,draw opacity=1 ][fill={rgb, 255:red, 144; green, 19; blue, 254 }  ,fill opacity=1 ][line width=1.5]      (0, 0) circle [x radius= 2.61, y radius= 2.61]   ;
\draw [color={rgb, 255:red, 144; green, 19; blue, 254 }  ,draw opacity=1 ][line width=1.5]  [dash pattern={on 1.69pt off 2.76pt}]  (395.75,436.5) -- (284.75,434) ;
\draw [color={rgb, 255:red, 172; green, 150; blue, 107 }  ,draw opacity=1 ][line width=1.5]  [dash pattern={on 1.69pt off 2.76pt}]  (381,420) -- (305,415) ;
\draw [color={rgb, 255:red, 142; green, 115; blue, 163 }  ,draw opacity=1 ][line width=1.5]  [dash pattern={on 1.69pt off 2.76pt}]  (366,365) -- (306.75,377.5) ;
\draw    (265.76,402.86) -- (299.74,380.64) ;
\draw [shift={(302.25,379)}, rotate = 146.82] [fill={rgb, 255:red, 0; green, 0; blue, 0 }  ][line width=0.08]  [draw opacity=0] (8.04,-3.86) -- (0,0) -- (8.04,3.86) -- (5.34,0) -- cycle    ;
\draw [shift={(263.25,404.5)}, rotate = 326.82] [fill={rgb, 255:red, 0; green, 0; blue, 0 }  ][line width=0.08]  [draw opacity=0] (8.04,-3.86) -- (0,0) -- (8.04,3.86) -- (5.34,0) -- cycle    ;
\draw    (268.7,408.55) -- (297.8,413.95) ;
\draw [shift={(300.75,414.5)}, rotate = 190.52] [fill={rgb, 255:red, 0; green, 0; blue, 0 }  ][line width=0.08]  [draw opacity=0] (8.04,-3.86) -- (0,0) -- (8.04,3.86) -- (5.34,0) -- cycle    ;
\draw [shift={(265.75,408)}, rotate = 10.52] [fill={rgb, 255:red, 0; green, 0; blue, 0 }  ][line width=0.08]  [draw opacity=0] (8.04,-3.86) -- (0,0) -- (8.04,3.86) -- (5.34,0) -- cycle    ;
\draw    (265,415) -- (269.72,420.98) ;
\draw [shift={(271.58,423.33)}, rotate = 231.69] [fill={rgb, 255:red, 0; green, 0; blue, 0 }  ][line width=0.08]  [draw opacity=0] (8.04,-3.86) -- (0,0) -- (8.04,3.86) -- (5.34,0) -- cycle    ;
\draw    (273.7,425.45) -- (279.08,430.83) ;
\draw [shift={(271.58,423.33)}, rotate = 45] [fill={rgb, 255:red, 0; green, 0; blue, 0 }  ][line width=0.08]  [draw opacity=0] (8.04,-3.86) -- (0,0) -- (8.04,3.86) -- (5.34,0) -- cycle    ;
\draw  [color={rgb, 255:red, 126; green, 211; blue, 33 }  ,draw opacity=1 ][line width=1.5]  (95,388) .. controls (95,359.28) and (118.28,336) .. (147,336) .. controls (175.72,336) and (199,359.28) .. (199,388) .. controls (199,416.72) and (175.72,440) .. (147,440) .. controls (118.28,440) and (95,416.72) .. (95,388) -- cycle ;
\draw [color={rgb, 255:red, 126; green, 211; blue, 33 }  ,draw opacity=1 ][line width=1.5]  [dash pattern={on 1.69pt off 2.76pt}]  (195,370) -- (250,350) ;
\draw [shift={(250,350)}, rotate = 340.02] [color={rgb, 255:red, 126; green, 211; blue, 33 }  ,draw opacity=1 ][fill={rgb, 255:red, 126; green, 211; blue, 33 }  ,fill opacity=1 ][line width=1.5]      (0, 0) circle [x radius= 2.61, y radius= 2.61]   ;
\draw [color={rgb, 255:red, 74; green, 108; blue, 134 }  ,draw opacity=1 ][line width=1.5]  [dash pattern={on 1.69pt off 2.76pt}]  (350,350) -- (288.25,348) ;
\draw [color={rgb, 255:red, 126; green, 211; blue, 33 }  ,draw opacity=1 ][line width=1.5]  [dash pattern={on 1.69pt off 2.76pt}]  (364.25,410) -- (261.25,381) ;
\draw    (259.25,348.34) -- (280.75,347.16) ;
\draw [shift={(283.75,347)}, rotate = 176.88] [fill={rgb, 255:red, 0; green, 0; blue, 0 }  ][line width=0.08]  [draw opacity=0] (8.04,-3.86) -- (0,0) -- (8.04,3.86) -- (5.34,0) -- cycle    ;
\draw [shift={(256.25,348.5)}, rotate = 356.88] [fill={rgb, 255:red, 0; green, 0; blue, 0 }  ][line width=0.08]  [draw opacity=0] (8.04,-3.86) -- (0,0) -- (8.04,3.86) -- (5.34,0) -- cycle    ;
\draw    (252.52,355.92) -- (254.63,363.25) ;
\draw [shift={(255.46,366.13)}, rotate = 253.93] [fill={rgb, 255:red, 0; green, 0; blue, 0 }  ][line width=0.08]  [draw opacity=0] (8.04,-3.86) -- (0,0) -- (8.04,3.86) -- (5.34,0) -- cycle    ;
\draw    (256.62,368.9) -- (259.56,375.91) ;
\draw [shift={(255.46,366.13)}, rotate = 67.24] [fill={rgb, 255:red, 0; green, 0; blue, 0 }  ][line width=0.08]  [draw opacity=0] (8.04,-3.86) -- (0,0) -- (8.04,3.86) -- (5.34,0) -- cycle    ;

\draw  [color={rgb, 255:red, 142; green, 115; blue, 163 }  ,draw opacity=1 ][line width=1.5]  (366,357) -- (382,357) -- (382,373) -- (366,373) -- cycle ;
\draw  [color={rgb, 255:red, 142; green, 115; blue, 163 }  ,draw opacity=1 ][fill={rgb, 255:red, 142; green, 115; blue, 163 }  ,fill opacity=1 ] (304.25,375) -- (309.25,375) -- (309.25,380) -- (304.25,380) -- cycle ;
\draw  [color={rgb, 255:red, 74; green, 108; blue, 134 }  ,draw opacity=1 ][fill={rgb, 255:red, 74; green, 108; blue, 134 }  ,fill opacity=1 ] (285.75,345.5) -- (290.75,345.5) -- (290.75,350.5) -- (285.75,350.5) -- cycle ;
\draw  [color={rgb, 255:red, 172; green, 150; blue, 107 }  ,draw opacity=1 ][fill={rgb, 255:red, 172; green, 150; blue, 107 }  ,fill opacity=1 ] (302.5,412.5) -- (307.5,412.5) -- (307.5,417.5) -- (302.5,417.5) -- cycle ;
\draw  [color={rgb, 255:red, 144; green, 19; blue, 254 }  ,draw opacity=1 ][fill={rgb, 255:red, 144; green, 19; blue, 254 }  ,fill opacity=1 ] (282.25,431.5) -- (287.25,431.5) -- (287.25,436.5) -- (282.25,436.5) -- cycle ;
\draw  [color={rgb, 255:red, 126; green, 211; blue, 33 }  ,draw opacity=1 ][fill={rgb, 255:red, 126; green, 211; blue, 33 }  ,fill opacity=1 ] (258.75,378.5) -- (263.75,378.5) -- (263.75,383.5) -- (258.75,383.5) -- cycle ;

\draw (226,467) node [anchor=north west][inner sep=0.75pt]  [font=\scriptsize] [align=left] {Shared Latent Space};
\draw (118,467) node [anchor=north west][inner sep=0.75pt]  [font=\scriptsize] [align=left] {Point Clouds};
\draw (368,467) node [anchor=north west][inner sep=0.75pt]  [font=\scriptsize] [align=left] {Renderings};

\end{tikzpicture}

%% file: tikz/architecture.tex
\graphicspath{{./figures/architecture/}} 


\tikzset{
	pattern size/.store in=\mcSize, 
	pattern size = 5pt,
	pattern thickness/.store in=\mcThickness, 
	pattern thickness = 0.3pt,
	pattern radius/.store in=\mcRadius, 
	pattern radius = 1pt}
\makeatletter
\pgfutil@ifundefined{pgf@pattern@name@_bswssf3zn}{
	\pgfdeclarepatternformonly[\mcThickness,\mcSize]{_bswssf3zn}
	{\pgfqpoint{0pt}{0pt}}
	{\pgfpoint{\mcSize+\mcThickness}{\mcSize+\mcThickness}}
	{\pgfpoint{\mcSize}{\mcSize}}
	{
		\pgfsetcolor{\tikz@pattern@color}
		\pgfsetlinewidth{\mcThickness}
		\pgfpathmoveto{\pgfqpoint{0pt}{0pt}}
		\pgfpathlineto{\pgfpoint{\mcSize+\mcThickness}{\mcSize+\mcThickness}}
		\pgfusepath{stroke}
}}
\makeatother
\tikzset{every picture/.style={line width=0.75pt}} 

\begin{tikzpicture}[x=0.75pt,y=0.75pt,yscale=-1,xscale=1]

\draw  [color={rgb, 255:red, 179; green, 176; blue, 176 }  ,draw opacity=1 ][fill={rgb, 255:red, 195; green, 236; blue, 243 }  ,fill opacity=1 ] (135,167) .. controls (135,160.37) and (140.37,155) .. (147,155) -- (243,155) .. controls (249.63,155) and (255,160.37) .. (255,167) -- (255,203) .. controls (255,209.63) and (249.63,215) .. (243,215) -- (147,215) .. controls (140.37,215) and (135,209.63) .. (135,203) -- cycle ;
\draw  [color={rgb, 255:red, 179; green, 176; blue, 176 }  ,draw opacity=1 ][pattern=_bswssf3zn,pattern size=6pt,pattern thickness=2.25pt,pattern radius=0pt, pattern color={rgb, 255:red, 255; green, 255; blue, 255}] (135,167) .. controls (135,160.37) and (140.37,155) .. (147,155) -- (243,155) .. controls (249.63,155) and (255,160.37) .. (255,167) -- (255,203) .. controls (255,209.63) and (249.63,215) .. (243,215) -- (147,215) .. controls (140.37,215) and (135,209.63) .. (135,203) -- cycle ;
\draw  [color={rgb, 255:red, 179; green, 176; blue, 176 }  ,draw opacity=1 ][fill={rgb, 255:red, 195; green, 236; blue, 243 }  ,fill opacity=1 ] (135,52) .. controls (135,45.37) and (140.37,40) .. (147,40) -- (243,40) .. controls (249.63,40) and (255,45.37) .. (255,52) -- (255,88) .. controls (255,94.63) and (249.63,100) .. (243,100) -- (147,100) .. controls (140.37,100) and (135,94.63) .. (135,88) -- cycle ;
\draw    (115,125) -- (182,125) ;
\draw [shift={(185,125)}, rotate = 180] [fill={rgb, 255:red, 0; green, 0; blue, 0 }  ][line width=0.08]  [draw opacity=0] (6.25,-3) -- (0,0) -- (6.25,3) -- cycle    ;
\draw    (110,70) -- (132,70) ;
\draw [shift={(135,70)}, rotate = 180] [fill={rgb, 255:red, 0; green, 0; blue, 0 }  ][line width=0.08]  [draw opacity=0] (6.25,-3) -- (0,0) -- (6.25,3) -- cycle    ;
\draw  [line width=0.75]  (70.84,149.37) -- (92.08,133.28) -- (104.02,142) -- (95.14,167.12) -- cycle ;
\draw    (110,185) -- (132,185) ;
\draw [shift={(135,185)}, rotate = 180] [fill={rgb, 255:red, 0; green, 0; blue, 0 }  ][line width=0.08]  [draw opacity=0] (6.25,-3) -- (0,0) -- (6.25,3) -- cycle    ;
\draw    (285,125) -- (285,103) ;
\draw [shift={(285,100)}, rotate = 90] [fill={rgb, 255:red, 0; green, 0; blue, 0 }  ][line width=0.08]  [draw opacity=0] (6.25,-3) -- (0,0) -- (6.25,3) -- cycle    ;
\draw (103.86,129.24) node [rotate=-36.62] {\includegraphics[width=19.4pt,height=14.49pt]{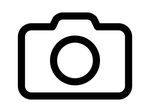}};
\draw  [fill={rgb, 255:red, 255; green, 255; blue, 255 }  ,fill opacity=1 ] (235,53) -- (245,53) -- (245,87) -- (235,87) -- cycle ;
\draw  [fill={rgb, 255:red, 255; green, 255; blue, 255 }  ,fill opacity=1 ] (175,55) -- (210,55) -- (210,63) -- (175,63) -- cycle ;
\draw  [fill={rgb, 255:red, 255; green, 255; blue, 255 }  ,fill opacity=1 ] (175,66) -- (210,66) -- (210,74) -- (175,74) -- cycle ;
\draw  [fill={rgb, 255:red, 255; green, 255; blue, 255 }  ,fill opacity=1 ] (175,77) -- (210,77) -- (210,85) -- (175,85) -- cycle ;
\draw    (155,70) -- (167,70) ;
\draw [shift={(170,70)}, rotate = 180] [fill={rgb, 255:red, 0; green, 0; blue, 0 }  ][line width=0.08]  [draw opacity=0] (6.25,-3) -- (0,0) -- (6.25,3) -- cycle    ;
\draw    (245,60) -- (272,60) ;
\draw [shift={(275,60)}, rotate = 180] [fill={rgb, 255:red, 0; green, 0; blue, 0 }  ][line width=0.08]  [draw opacity=0] (6.25,-3) -- (0,0) -- (6.25,3) -- cycle    ;
\draw    (195,90) -- (272,90) ;
\draw [shift={(275,90)}, rotate = 180] [fill={rgb, 255:red, 0; green, 0; blue, 0 }  ][line width=0.08]  [draw opacity=0] (6.25,-3) -- (0,0) -- (6.25,3) -- cycle    ;
\draw    (195,85) -- (195,90) ;
\draw  [fill={rgb, 255:red, 255; green, 255; blue, 255 }  ,fill opacity=1 ] (235,168) -- (245,168) -- (245,202) -- (235,202) -- cycle ;
\draw  [fill={rgb, 255:red, 255; green, 255; blue, 255 }  ,fill opacity=1 ] (175,170) -- (210,170) -- (210,178) -- (175,178) -- cycle ;
\draw  [fill={rgb, 255:red, 255; green, 255; blue, 255 }  ,fill opacity=1 ] (175,180.77) -- (210,180.77) -- (210,188.77) -- (175,188.77) -- cycle ;
\draw  [fill={rgb, 255:red, 255; green, 255; blue, 255 }  ,fill opacity=1 ] (175,191.77) -- (210,191.77) -- (210,199.77) -- (175,199.77) -- cycle ;
\draw    (155,185) -- (167,185) ;
\draw [shift={(170,185)}, rotate = 180] [fill={rgb, 255:red, 0; green, 0; blue, 0 }  ][line width=0.08]  [draw opacity=0] (6.25,-3) -- (0,0) -- (6.25,3) -- cycle    ;
\draw    (195,200) -- (195,205) ;
\draw (58.34,73.02) node  {\includegraphics[width=72.51pt,height=49.53pt]{transparent.png}};
\draw (60.59,179) node  {\includegraphics[width=63.64pt,height=54pt]{model_normalized_003_cropped.png}};
\draw  [fill={rgb, 255:red, 255; green, 255; blue, 255 }  ,fill opacity=1 ] (218,85) -- (218,55) -- (228,70) -- cycle ;
\draw  [fill={rgb, 255:red, 255; green, 255; blue, 255 }  ,fill opacity=1 ] (218,200) -- (218,170) -- (228,185) -- cycle ;
\draw  [color={rgb, 255:red, 0; green, 0; blue, 0 }  ,draw opacity=1 ][dash pattern={on 4.5pt off 4.5pt}] (330,11) -- (580,11) -- (580,250) -- (330,250) -- cycle ;
\draw  [color={rgb, 255:red, 126; green, 211; blue, 33 }  ,draw opacity=1 ][line width=1.5]  (402.84,70) .. controls (402.84,41.28) and (426.12,18) .. (454.84,18) .. controls (483.56,18) and (506.84,41.28) .. (506.84,70) .. controls (506.84,98.72) and (483.56,122) .. (454.84,122) .. controls (426.12,122) and (402.84,98.72) .. (402.84,70) -- cycle ;
\draw  [color={rgb, 255:red, 255; green, 255; blue, 255 }  ,draw opacity=1 ][fill={rgb, 255:red, 255; green, 255; blue, 255 }  ,fill opacity=1 ] (480,145) -- (570,145) -- (570,235) -- (480,235) -- cycle ;
\draw (526.24,187.75) node  {\includegraphics[width=62.62pt,height=50.63pt]{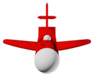}};
\draw  [color={rgb, 255:red, 155; green, 155; blue, 155 }  ,draw opacity=1 ] (480,145) -- (570,145) -- (570,235) -- (480,235) -- cycle ;
\draw [color={rgb, 255:red, 155; green, 155; blue, 155 }  ,draw opacity=1 ]   (498,145) -- (498,235) ;
\draw [color={rgb, 255:red, 155; green, 155; blue, 155 }  ,draw opacity=1 ]   (516,145) -- (516,235) ;
\draw [color={rgb, 255:red, 155; green, 155; blue, 155 }  ,draw opacity=1 ]   (534,145) -- (534,235) ;
\draw [color={rgb, 255:red, 155; green, 155; blue, 155 }  ,draw opacity=1 ]   (552,145) -- (552,235) ;
\draw [color={rgb, 255:red, 155; green, 155; blue, 155 }  ,draw opacity=1 ]   (480,163) -- (570,163) ;
\draw [color={rgb, 255:red, 155; green, 155; blue, 155 }  ,draw opacity=1 ]   (480,181) -- (570,181) ;
\draw [color={rgb, 255:red, 155; green, 155; blue, 155 }  ,draw opacity=1 ]   (480,199) -- (570,199) ;
\draw [color={rgb, 255:red, 155; green, 155; blue, 155 }  ,draw opacity=1 ]   (480,217) -- (570,217) ;

\draw  [color={rgb, 255:red, 255; green, 255; blue, 255 }  ,draw opacity=1 ][fill={rgb, 255:red, 255; green, 255; blue, 255 }  ,fill opacity=1 ] (410,150) -- (500,150) -- (500,240) -- (410,240) -- cycle ;
\draw (456.11,197.25) node  {\includegraphics[width=65.83pt,height=57.38pt]{model_normalized_007_cropped.png}};
\draw  [color={rgb, 255:red, 155; green, 155; blue, 155 }  ,draw opacity=1 ] (410,150) -- (500,150) -- (500,240) -- (410,240) -- cycle ;
\draw [color={rgb, 255:red, 155; green, 155; blue, 155 }  ,draw opacity=1 ]   (428,150) -- (428,240) ;
\draw [color={rgb, 255:red, 155; green, 155; blue, 155 }  ,draw opacity=1 ]   (446,150) -- (446,240) ;
\draw [color={rgb, 255:red, 155; green, 155; blue, 155 }  ,draw opacity=1 ]   (464,150) -- (464,240) ;
\draw [color={rgb, 255:red, 155; green, 155; blue, 155 }  ,draw opacity=1 ]   (482,150) -- (482,240) ;
\draw [color={rgb, 255:red, 155; green, 155; blue, 155 }  ,draw opacity=1 ]   (410,168) -- (500,168) ;
\draw [color={rgb, 255:red, 155; green, 155; blue, 155 }  ,draw opacity=1 ]   (410,186) -- (500,186) ;
\draw [color={rgb, 255:red, 155; green, 155; blue, 155 }  ,draw opacity=1 ]   (410,204) -- (500,204) ;
\draw [color={rgb, 255:red, 155; green, 155; blue, 155 }  ,draw opacity=1 ]   (410,222) -- (500,222) ;

\draw  [color={rgb, 255:red, 255; green, 255; blue, 255 }  ,draw opacity=1 ][fill={rgb, 255:red, 255; green, 255; blue, 255 }  ,fill opacity=1 ] (340,155) -- (430,155) -- (430,245) -- (340,245) -- cycle ;
\draw (382.43,200) node  {\includegraphics[width=63.64pt,height=54pt]{model_normalized_003_cropped.png}};
\draw  [color={rgb, 255:red, 155; green, 155; blue, 155 }  ,draw opacity=1 ] (340,155) -- (430,155) -- (430,245) -- (340,245) -- cycle ;
\draw [color={rgb, 255:red, 155; green, 155; blue, 155 }  ,draw opacity=1 ]   (358,155) -- (358,245) ;
\draw [color={rgb, 255:red, 155; green, 155; blue, 155 }  ,draw opacity=1 ]   (376,155) -- (376,245) ;
\draw [color={rgb, 255:red, 155; green, 155; blue, 155 }  ,draw opacity=1 ]   (394,155) -- (394,245) ;
\draw [color={rgb, 255:red, 155; green, 155; blue, 155 }  ,draw opacity=1 ]   (412,155) -- (412,245) ;
\draw [color={rgb, 255:red, 155; green, 155; blue, 155 }  ,draw opacity=1 ]   (340,173) -- (430,173) ;
\draw [color={rgb, 255:red, 155; green, 155; blue, 155 }  ,draw opacity=1 ]   (340,191) -- (430,191) ;
\draw [color={rgb, 255:red, 155; green, 155; blue, 155 }  ,draw opacity=1 ]   (340,209) -- (430,209) ;
\draw [color={rgb, 255:red, 155; green, 155; blue, 155 }  ,draw opacity=1 ]   (340,227) -- (430,227) ;
\draw  [color={rgb, 255:red, 126; green, 211; blue, 33 }  ,draw opacity=1 ][line width=1.5]  (340,155) -- (430,155) -- (430,245) -- (340,245) -- cycle ;
\draw  [color={rgb, 255:red, 245; green, 166; blue, 35 }  ,draw opacity=1 ][line width=1.5]  (412,191) -- (430,191) -- (430,209) -- (412,209) -- cycle ;
\draw  [color={rgb, 255:red, 144; green, 19; blue, 254 }  ,draw opacity=1 ][line width=1.5]  (376,209) -- (394,209) -- (394,227) -- (376,227) -- cycle ;
\draw [color={rgb, 255:red, 255; green, 0; blue, 0 }  ,draw opacity=1 ][line width=1.5]  [dash pattern={on 1.69pt off 2.76pt}]  (430.07,90.26) -- (352.5,186.88) ;
\draw [shift={(350,190)}, rotate = 308.76] [fill={rgb, 255:red, 255; green, 0; blue, 0 }  ,fill opacity=1 ][line width=0.08]  [draw opacity=0] (8.13,-3.9) -- (0,0) -- (8.13,3.9) -- cycle    ;
\draw [color={rgb, 255:red, 144; green, 19; blue, 254 }  ,draw opacity=1 ][line width=1.5]  [dash pattern={on 1.69pt off 2.76pt}]  (460.91,99.41) -- (387.28,205.71) ;
\draw [shift={(385,209)}, rotate = 304.71] [fill={rgb, 255:red, 144; green, 19; blue, 254 }  ,fill opacity=1 ][line width=0.08]  [draw opacity=0] (8.13,-3.9) -- (0,0) -- (8.13,3.9) -- cycle    ;
\draw [color={rgb, 255:red, 126; green, 211; blue, 33 }  ,draw opacity=1 ][line width=1.5]  [dash pattern={on 1.69pt off 2.76pt}]  (425,115) -- (397.4,151.8) ;
\draw [shift={(395,155)}, rotate = 306.87] [fill={rgb, 255:red, 126; green, 211; blue, 33 }  ,fill opacity=1 ][line width=0.08]  [draw opacity=0] (8.13,-3.9) -- (0,0) -- (8.13,3.9) -- cycle    ;
\draw (452.18,76.98) node  {\includegraphics[width=72.51pt,height=49.53pt]{transparent.png}};
\draw  [color={rgb, 255:red, 144; green, 19; blue, 254 }  ,draw opacity=1 ][line width=1.5]  (455.98,99.41) .. controls (455.98,96.69) and (458.18,94.48) .. (460.91,94.48) .. controls (463.63,94.48) and (465.84,96.69) .. (465.84,99.41) .. controls (465.84,102.13) and (463.63,104.33) .. (460.91,104.33) .. controls (458.18,104.33) and (455.98,102.13) .. (455.98,99.41) -- cycle ;
\draw  [color={rgb, 255:red, 255; green, 0; blue, 0 }  ,draw opacity=1 ][line width=1.5]  (425.13,90.26) .. controls (425.13,87.54) and (427.34,85.33) .. (430.07,85.33) .. controls (432.79,85.33) and (435,87.54) .. (435,90.26) .. controls (435,92.98) and (432.79,95.18) .. (430.07,95.18) .. controls (427.34,95.18) and (425.13,92.98) .. (425.13,90.26) -- cycle ;
\draw  [color={rgb, 255:red, 245; green, 166; blue, 35 }  ,draw opacity=1 ][line width=1.5]  (488.84,45.46) .. controls (488.84,42.96) and (490.88,40.92) .. (493.39,40.92) .. controls (495.9,40.92) and (497.93,42.96) .. (497.93,45.46) .. controls (497.93,47.97) and (495.9,50) .. (493.39,50) .. controls (490.88,50) and (488.84,47.97) .. (488.84,45.46) -- cycle ;
\draw [color={rgb, 255:red, 245; green, 166; blue, 35 }  ,draw opacity=1 ][line width=1.5]  [dash pattern={on 1.69pt off 2.76pt}]  (493.39,45.46) -- (421.33,186.44) ;
\draw [shift={(419.51,190)}, rotate = 297.07] [fill={rgb, 255:red, 245; green, 166; blue, 35 }  ,fill opacity=1 ][line width=0.08]  [draw opacity=0] (8.13,-3.9) -- (0,0) -- (8.13,3.9) -- cycle    ;
\draw  [color={rgb, 255:red, 255; green, 0; blue, 0 }  ,draw opacity=1 ][line width=1.5]  (340,191) -- (358,191) -- (358,209) -- (340,209) -- cycle ;
\draw  [color={rgb, 255:red, 179; green, 176; blue, 176 }  ,draw opacity=1 ][fill={rgb, 255:red, 253; green, 210; blue, 134 }  ,fill opacity=1 ] (275,54) .. controls (275,51.79) and (276.79,50) .. (279,50) -- (291,50) .. controls (293.21,50) and (295,51.79) .. (295,54) -- (295,66) .. controls (295,68.21) and (293.21,70) .. (291,70) -- (279,70) .. controls (276.79,70) and (275,68.21) .. (275,66) -- cycle ;
\draw  [color={rgb, 255:red, 179; green, 176; blue, 176 }  ,draw opacity=1 ][fill={rgb, 255:red, 253; green, 210; blue, 134 }  ,fill opacity=1 ] (275,84) .. controls (275,81.79) and (276.79,80) .. (279,80) -- (291,80) .. controls (293.21,80) and (295,81.79) .. (295,84) -- (295,96) .. controls (295,98.21) and (293.21,100) .. (291,100) -- (279,100) .. controls (276.79,100) and (275,98.21) .. (275,96) -- cycle ;
\draw  [color={rgb, 255:red, 179; green, 176; blue, 176 }  ,draw opacity=1 ][fill={rgb, 255:red, 202; green, 238; blue, 166 }  ,fill opacity=1 ] (185,119.5) .. controls (185,117.57) and (186.57,116) .. (188.5,116) -- (201.5,116) .. controls (203.43,116) and (205,117.57) .. (205,119.5) -- (205,130) .. controls (205,131.93) and (203.43,133.5) .. (201.5,133.5) -- (188.5,133.5) .. controls (186.57,133.5) and (185,131.93) .. (185,130) -- cycle ;
\draw  [draw opacity=0][fill={rgb, 255:red, 155; green, 155; blue, 155 }  ,fill opacity=1 ] (141.21,165.32) .. controls (141.21,164.78) and (141.65,164.33) .. (142.19,164.33) -- (149.01,164.33) .. controls (149.56,164.33) and (150,164.78) .. (150,165.32) -- (150,170.01) .. controls (150,170.56) and (149.56,171) .. (149.01,171) -- (142.19,171) .. controls (141.65,171) and (141.21,170.56) .. (141.21,170.01) -- cycle ;
\draw  [draw opacity=0][fill={rgb, 255:red, 155; green, 155; blue, 155 }  ,fill opacity=1 ] (142.16,166.24) -- (142.16,163.38) .. controls (142.16,161.52) and (143.67,160) .. (145.54,160) -- (145.54,160) .. controls (147.41,160) and (148.93,161.52) .. (148.93,163.38) -- (148.93,166.24) -- (148.94,166.24) -- (148.39,166.24) -- (147.84,166.24) -- (147.84,166.24) -- (147.84,163.38) .. controls (147.84,162.11) and (146.81,161.08) .. (145.54,161.08) -- (145.54,161.08) .. controls (144.27,161.08) and (143.24,162.11) .. (143.24,163.38) -- (143.24,166.24) -- cycle ;

\draw    (245,175) -- (272,175) ;
\draw [shift={(275,175)}, rotate = 180] [fill={rgb, 255:red, 0; green, 0; blue, 0 }  ][line width=0.08]  [draw opacity=0] (6.25,-3) -- (0,0) -- (6.25,3) -- cycle    ;
\draw    (294,205) -- (336,205) ;
\draw [shift={(339,205)}, rotate = 180] [fill={rgb, 255:red, 0; green, 0; blue, 0 }  ][line width=0.08]  [draw opacity=0] (6.25,-3) -- (0,0) -- (6.25,3) -- cycle    ;
\draw  [color={rgb, 255:red, 179; green, 176; blue, 176 }  ,draw opacity=1 ][fill={rgb, 255:red, 253; green, 210; blue, 134 }  ,fill opacity=1 ] (275,169) .. controls (275,166.79) and (276.79,165) .. (279,165) -- (291,165) .. controls (293.21,165) and (295,166.79) .. (295,169) -- (295,181) .. controls (295,183.21) and (293.21,185) .. (291,185) -- (279,185) .. controls (276.79,185) and (275,183.21) .. (275,181) -- cycle ;
\draw  [color={rgb, 255:red, 179; green, 176; blue, 176 }  ,draw opacity=1 ][fill={rgb, 255:red, 253; green, 210; blue, 134 }  ,fill opacity=1 ] (275,199) .. controls (275,196.79) and (276.79,195) .. (279,195) -- (291,195) .. controls (293.21,195) and (295,196.79) .. (295,199) -- (295,211) .. controls (295,213.21) and (293.21,215) .. (291,215) -- (279,215) .. controls (276.79,215) and (275,213.21) .. (275,211) -- cycle ;
\draw    (295,175) -- (336,175) ;
\draw [shift={(339,175)}, rotate = 180] [fill={rgb, 255:red, 0; green, 0; blue, 0 }  ][line width=0.08]  [draw opacity=0] (6.25,-3) -- (0,0) -- (6.25,3) -- cycle    ;
\draw    (195,205) -- (272,205.22) ;
\draw [shift={(275,205.23)}, rotate = 180.16] [fill={rgb, 255:red, 0; green, 0; blue, 0 }  ][line width=0.08]  [draw opacity=0] (6.25,-3) -- (0,0) -- (6.25,3) -- cycle    ;
\draw    (295,60) -- (400,60) ;
\draw [shift={(403,60)}, rotate = 180] [fill={rgb, 255:red, 0; green, 0; blue, 0 }  ][line width=0.08]  [draw opacity=0] (6.25,-3) -- (0,0) -- (6.25,3) -- cycle    ;
\draw    (295,90) -- (421,90) ;
\draw [shift={(424,90)}, rotate = 180] [fill={rgb, 255:red, 0; green, 0; blue, 0 }  ][line width=0.08]  [draw opacity=0] (6.25,-3) -- (0,0) -- (6.25,3) -- cycle    ;
\draw    (205,125) -- (285,125) ;

\draw (160,104) node [anchor=north west][inner sep=0.75pt]  [font=\scriptsize] [align=left] {Pose Encoder};
\draw (164,28) node [anchor=north west][inner sep=0.75pt]  [font=\scriptsize] [align=left] {3D Backbone};
\draw (141,70) node [anchor=north west][inner sep=0.75pt]  [font=\scriptsize] [align=left] {...};
\draw (146,143) node [anchor=north west][inner sep=0.75pt]  [font=\scriptsize] [align=left] {Frozen 2D Backbone};
\draw (141,185) node [anchor=north west][inner sep=0.75pt]  [font=\scriptsize] [align=left] {...};
\draw (261,17) node [anchor=north west][inner sep=0.75pt]  [font=\scriptsize] [align=left] {Projection\\Heads $\displaystyle \mathcal{H}$};
\draw (218,44) node [anchor=north west][inner sep=0.75pt]  [font=\scriptsize] [align=left] {$\displaystyle \mathcal{A}$};
\draw (218,159) node [anchor=north west][inner sep=0.75pt]  [font=\scriptsize] [align=left] {$\displaystyle \mathcal{A}$};
\draw (161,44) node [anchor=north west][inner sep=0.75pt]  [font=\scriptsize] [align=left] {$\displaystyle f^{\text{3d}}$};
\draw (161,159) node [anchor=north west][inner sep=0.75pt]  [font=\scriptsize] [align=left] {$\displaystyle f^{\text{2d}}$};

\end{tikzpicture}

%% file: tikz/heatmap.tex
\graphicspath{{./figures/visualization/}} 

\tikzset{every picture/.style={line width=0.75pt}} 

\begin{tikzpicture}[x=0.75pt,y=0.75pt,yscale=-1,xscale=1]

\draw (245.5,90.5) node  {\includegraphics[width=74.25pt,height=74.25pt]{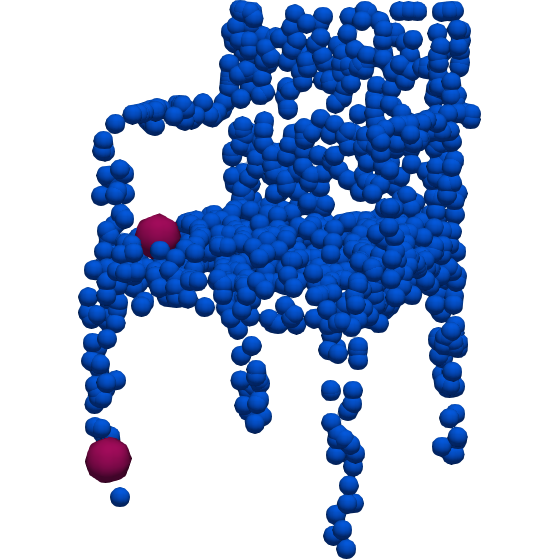}};
\draw (350,90) node  {\includegraphics[width=75pt,height=75pt]{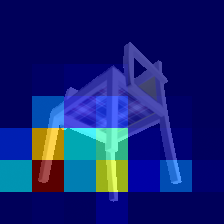}};
\draw (455,90) node  {\includegraphics[width=75pt,height=75pt]{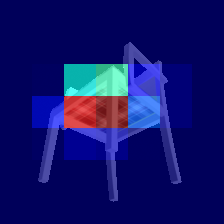}};
\draw  [draw opacity=0] (400,40) -- (400,245) -- (300,245) -- (300,40) -- cycle ;
\draw  [draw opacity=0] (505,40) -- (505,245) -- (405,245) -- (405,40) -- cycle ;
\draw  [draw opacity=0] (295,40) -- (295,245) -- (195,245) -- (195,40) -- cycle ;
\draw  [draw opacity=0] (400,145) -- (400,350) -- (300,350) -- (300,145) -- cycle ;
\draw  [draw opacity=0] (505,145) -- (505,350) -- (405,350) -- (405,145) -- cycle ;
\draw (455,195) node  {\includegraphics[width=75pt,height=75pt]{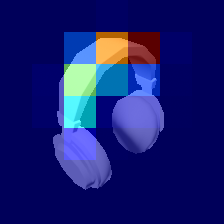}};
\draw (245,195) node  {\includegraphics[width=75pt,height=75pt]{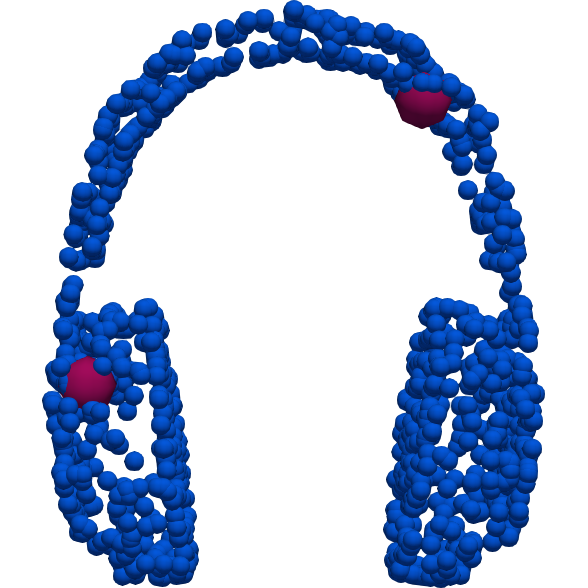}};
\draw (350,195) node  {\includegraphics[width=75pt,height=75pt]{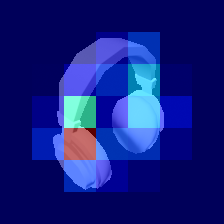}};
\draw (455,300) node  {\includegraphics[width=75pt,height=75pt]{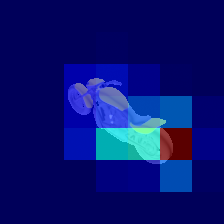}};
\draw (350,300) node  {\includegraphics[width=75pt,height=75pt]{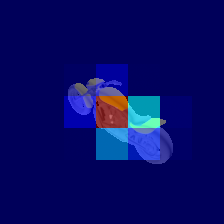}};
\draw (245,300) node  {\includegraphics[width=75pt,height=75pt]{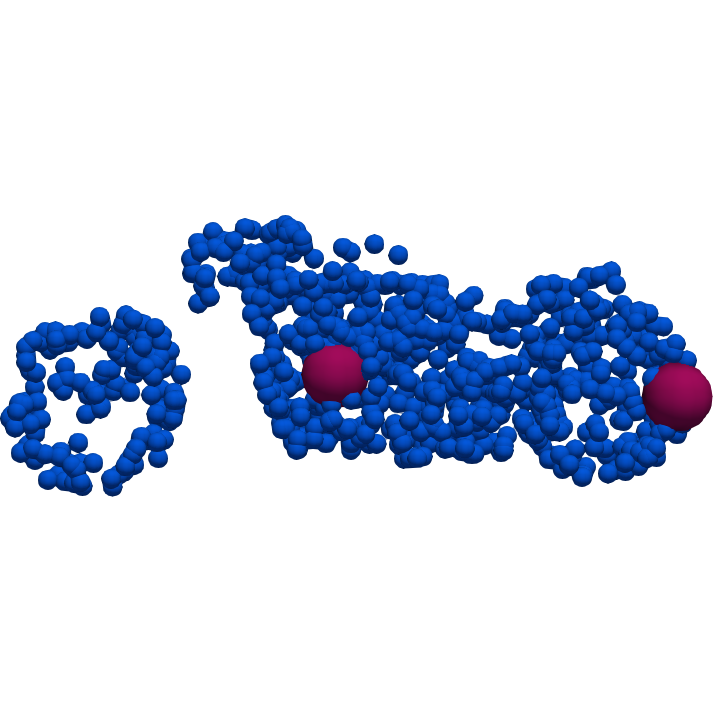}};

\draw (198,116) node [anchor=north west][inner sep=0.75pt]   [align=left] {\textbf{a}};
\draw (198.5,68) node [anchor=north west][inner sep=0.75pt]   [align=left] {\textbf{b}};
\draw (306,46) node [anchor=north west][inner sep=0.75pt]  [color={rgb, 255:red, 255; green, 255; blue, 255 }  ,opacity=1 ] [align=left] {\textbf{a}};
\draw (411,46) node [anchor=north west][inner sep=0.75pt]  [color={rgb, 255:red, 255; green, 255; blue, 255 }  ,opacity=1 ] [align=left] {\textbf{b}};
\draw (306,152) node [anchor=north west][inner sep=0.75pt]  [color={rgb, 255:red, 255; green, 255; blue, 255 }  ,opacity=1 ] [align=left] {\textbf{c}};
\draw (411,151) node [anchor=north west][inner sep=0.75pt]  [color={rgb, 255:red, 255; green, 255; blue, 255 }  ,opacity=1 ] [align=left] {\textbf{d}};
\draw (232.5,318) node [anchor=north west][inner sep=0.75pt]   [align=left] {\textbf{e}};
\draw (306,256) node [anchor=north west][inner sep=0.75pt]  [color={rgb, 255:red, 255; green, 255; blue, 255 }  ,opacity=1 ] [align=left] {\textbf{e}};
\draw (282,320.5) node [anchor=north west][inner sep=0.75pt]   [align=left] {\textbf{f}};
\draw (411,256) node [anchor=north west][inner sep=0.75pt]  [color={rgb, 255:red, 255; green, 255; blue, 255 }  ,opacity=1 ] [align=left] {\textbf{f}};
\draw (228,192) node [anchor=north west][inner sep=0.75pt]   [align=left] {\textbf{c}};
\draw (251,167) node [anchor=north west][inner sep=0.75pt]   [align=left] {\textbf{d}};

\end{tikzpicture}